\documentclass[final]{cvpr}

\usepackage{times}
\usepackage{epsfig}
\usepackage{graphicx}
\usepackage{amsmath}
\usepackage{amssymb}
\usepackage{multirow}
\usepackage{color}
\usepackage{threeparttable}
\usepackage{soul}

\usepackage{todo}

\usepackage[pagebackref=true,breaklinks=true,colorlinks,bookmarks=false]{hyperref}

\def\cvprPaperID{10111} 
\def\confYear{CVPR 2021}

\begin{document}

\title{Learning Transferable 3D Adversarial Cloaks for Deep Trained Detectors}

\author{Arman Maesumi\\
University of Texas at Austin\\
{\tt\small arman@cs.utexas.edu}
\and
Mingkang Zhu\\
University of Texas at Austin\\
{\tt\small mz8374@utexas.edu}
\and
Yi Wang \\
University of Texas at Austin\\
{\tt\small panzer.wy@utexas.edu}
\and 
Tianlong Chen\\
University of Texas at Austin\\
{\tt\small tianlong.chen@utexas.edu}
\and 
Zhangyang Wang\\
University of Texas at Austin\\
{\tt\small atlaswang@utexas.edu}
\and 
Chandrajit Bajaj \\
University of Texas at Austin\\
{\tt\small bajaj@cs.utexas.edu}
}

\maketitle

\begin{abstract}
This paper presents a novel patch-based adversarial attack pipeline that trains adversarial patches on 3D human meshes. We sample triangular faces on a reference human mesh, and create an adversarial texture atlas over those faces. The adversarial texture is transferred to human meshes in various poses, which are rendered onto a collection of real-world background images. Contrary to the traditional patch-based adversarial attacks, where prior work attempts to fool trained object detectors using appended adversarial patches, this new form of attack is mapped into the 3D object world and back-propagated to the texture atlas through differentiable rendering. As such, the adversarial patch is trained under deformation consistent with real-world materials. In addition, and unlike existing adversarial patches, our new 3D adversarial patch is shown to fool state-of-the-art deep object detectors robustly under varying views, potentially leading to an attacking scheme that is persistently strong in the physical world. Our code is available via github: \url{https://github.com/CVC-Lab/3D_ADV_Mesh_pytorch3d}

\end{abstract}

\section{Introduction}

\begin{figure}[t]
\centering
\includegraphics[width=1\linewidth]{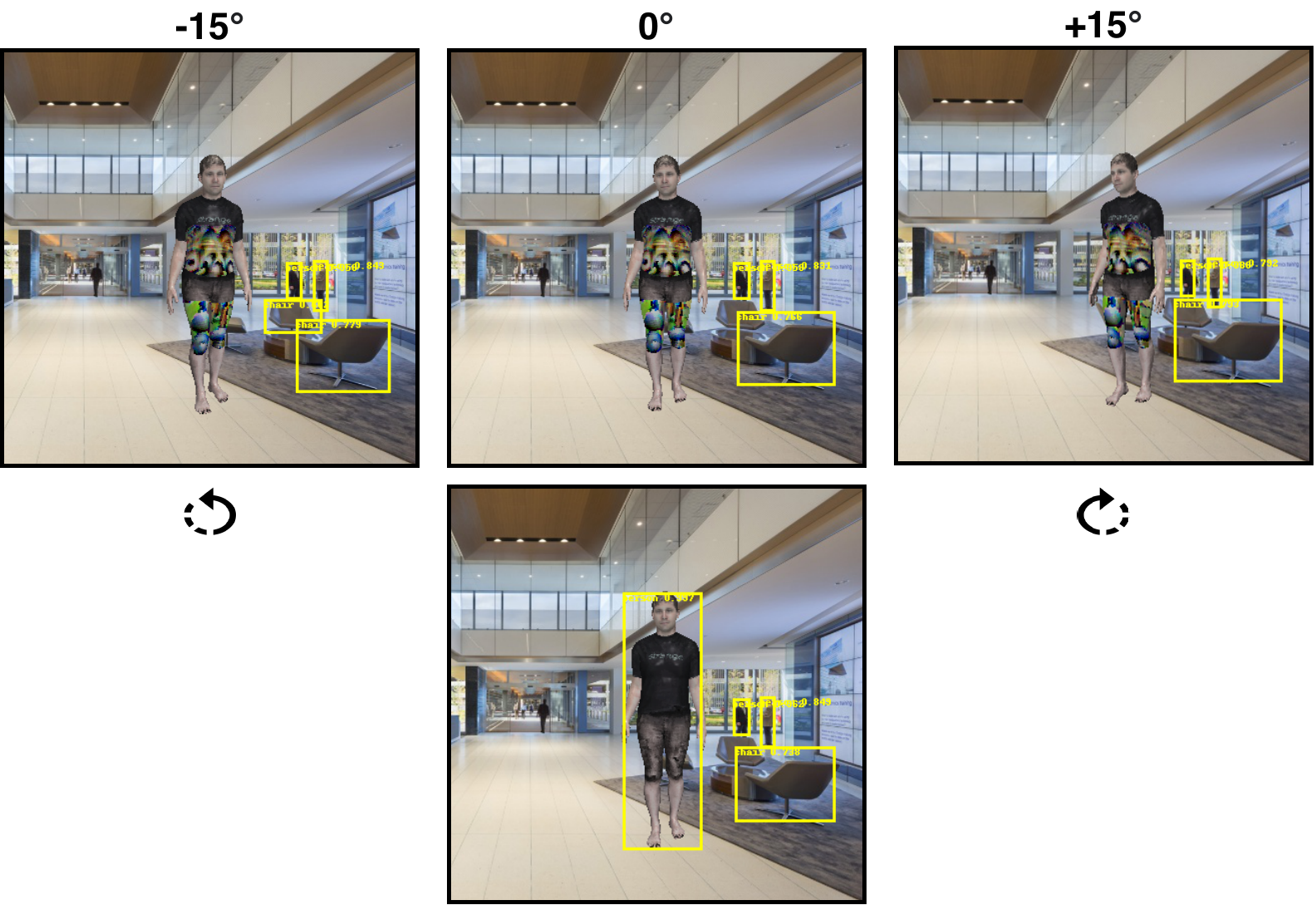}
\caption{An example of our 3D adversarial attack on a human mesh at different angles. The second row depicts the mesh without any adversarial perturbation; consequently, Faster R-CNN\cite{shaoqing2015fasterRCNN} identifies it as a human with $99\%$ confidence. The three adversarial images in row one are able to fool both Faster R-CNN and YoloV2\cite{redmon2017yolo9000}. Our 3D adversarial patch (on the chest and thighs) is viewed as part of the texture atlas over 3D human meshes.  When rendering 3D human meshes with varying poses, spatial locations, and camera angles, the attack remains robust, causing the mesh to be effectively cloaked.
}

\label{fig:intro}
\end{figure}


Deep neural networks are notoriously vulnerable to human-imperceivable perturbations or doctoring of images, resulting in the trained algorithms drastically changing their recognition and predictions. To test the misrecognition or misdetection vulnerability, Tram{\`e}r et al. \cite{tramer2017ensemble} propose 2D adversarial attacks, manipulating pixels on the image while maintaining overall visual fidelity. This negligible perturbation to human eyes causes drastically false conclusions with high confidence by trained deep neural networks. Numerous adversarial attacks have been designed and tested on deep learning tasks such as image classification and object detection. Among extensive efforts, the focus recently has shifted to only structurally editing certain local areas on an image, known as \textit{patch adversarial attacks} \cite{brown2017adversarial}. Thys et al.\cite{thys2019fooling} propose a pipeline to generate a 2D adversarial patch and attach it to image pixels of humans appearing in 2D images. In principle, a person with this 2D adversarial patch will fool or become ``invisible'' from deep learned human image detectors. However, such 2D image adversarial patches are often not robust to image transformations, especially under multi-view 2D image synthesis in reconstructed 3D computer graphics settings. Examining 2D image renderings from 3D scene models using various possible human postures and different viewing angles of humans, the 2D attack can easily lose its strength under such 3D viewing transformations. 
Moreover, while square or rectangular adversarial patches are typically under consideration, more shape variations and their implications for the attack performance have rarely been discussed before. 

Can we naturally stitch a patch onto human clothes to make the adversarial attack more versatile and realistic? The defect in pure 2D scenarios leads us to consider the 3D adversarial attack, where we view a person as a 3D object instead of its 2D projection. As an example, the domain of mesh adversarial attack \cite{xiao2019meshadv} refers to deformations in the mesh's shape and texture to fulfill the attack goal. However, these 3D adversarial attacks have not yet exemplified the concept of patch-based adversarial attacks; they view the entire texture and geometric information of 3D meshes as attackable. Moreover, a noticeable branch of research shows that 2D images with infinitesimal rotation and shift may cause huge perturbation in predictions \cite{zhang2019making,azulay2018deep,engstrom2019exploring}, no matter how negligible to human eyes. What if the perturbation does not come from 2D scenarios and conditions ($e.g.$, 2D rotation and translation), but rather results from changes in the physical world, like 3D view rotations and body postures changes? Furthermore, effective attacks on certain meshes do not imply a generalized effectiveness among other meshes. For instance, the attack can fail when the perturbations are applied to a mesh with different textures. Those downsides motivate us to develop a more generalized 3D adversarial patch. 

The primary aim of this work is to generate what we call a \textit{3D adversarial logo}, a structured patch in an arbitrary shape. When appended to a 3D human mesh, and rendered into 2D images, the logo should provide sufficient perturbation as to consistently fool object detectors, even under different human poses and viewing angles. A 3D adversarial logo is defined as a texture perturbation over a subregion of a mesh's given texture. Human meshes, along with 3D adversarial logos, are rendered and imposed on top of real-world background images. The specific contributions of our work are highlighted as:
\begin{itemize}


\item [$\bullet$] We propose a general 3D-to-2D adversarial attack protocol via physical rendering equipped with differentiability. With the 3D adversarial logo attached, we render 3D human meshes into 2D scenarios and synthesize images that fool object detectors. The shape of our 3D adversarial logo comes from sampled faces on our 3D human mesh. Hence, we can perform versatile adversarial training with various shapes and positions.

\item [$\bullet$] In order to create a more robust adversarial patch, we make use of the Skinned Multi-Person Linear Model (SMPL) \cite{SMPL:2015}, a generative model for the human body. We use the SMPL model to generate 3D human meshes in various poses, as to simulate more realistic imagery during training. Texture maps from the SURREAL Dataset \cite{varol17_surreal} are used on our 3D human meshes.

\item [$\bullet$] We justify that our model can adapt to multi-angle scenarios with much richer variations than what can be depicted by 2D perturbations, taking one important step towards studying the physical world fragility of deep networks. 

\end{itemize}

\section{Related Work}
\subsection{Differentiable Meshes}
Various tasks, including depth estimation as well as 3D reconstruction from 2D images, have been explored with deep neural networks and witnessed successes. Less considered is the reverse problem: How can we render the 3D model back to 2D images to fulfill desired tasks? 

Discrete operations in the two most popular rendering methods (ray-tracing and rasterization) hamper the differentiability. To fill in the gap, numerous approaches have been proposed to edit mesh texture via gradient descent, which provides the ground to combine traditional graphical renderer with neural networks. Nguyen-Phuoc et al. \cite{nguyen2018rendernet} propose a CNN architecture leveraging a projection unit to render a voxel-based 3D object into 2D images. Unlike the voxel-based method, Kato et al. \cite{kato2018neural} adopt linear-gradient interpolation to overcome vanishing gradients in rasterization-based rendering. Raj et  al. \cite{raj2019learning} generate textures for 3D mesh through photo-realistic pictures. They then apply RenderForCNN \cite{su2015render} to sample the viewpoints that match the ones of input images, followed by adapting CycleGAN \cite{zhu2017unpaired} to generate textures for 2.5D information rendered in the generated multi-viewpoints, and eventually merge these textures into a single texture to render the object into the 2D world.

\subsection{Adversarial Patches in 2D Images}
Adversarial attacks \cite{szegedy2013intriguing,goodfellow2014explaining,tkhu2019triplewins,Chen_2020_CVPR,gui2019ATMC} are proposed to analyze the robustness of CNNs, and recently are increasingly studied in object detection tasks, in the form of adversarial patches. For example, \cite{chen2018shapeshifter} provides a stop sign attack to Fast-RCNN \cite{girshick2015fast}, and \cite{thys2019fooling} is fooling the YOLOv2 \cite{redmon2017yolo9000} object detector through pixel-wise patch optimization. The target patch with simple 2D transformations (such as rotation and scaling) is applied to a near-human region in 2D real photos and then trained to fool with the object detector. To demonstrate realistic adversarial attacks, they physically let a person hold the 3D-printed patch and verify them to "disappear" in the object detector. Nevertheless, such attacks are easily broken w.r.t. real-world 3D variations as pointed out by \cite{lu2017no}. Wiyatno et al.  \cite{wiyatno2019physical} propose to generate physical adversarial texture as a patch in backgrounds. Their method allows the patch to be ``rotated'' in 3D space and then added back to 2D space. Xu et al.  \cite{xu2019evading} discusses how to incorporate physical deformation of T-shirts into patch adversarial attacks, leading a forward step yet only in a fixed camera view.  A recent work by Huang et al. \cite{huang2020universal} attacks region proposal networks (RPN) by synthesizing semantic patches that are naturally anchored onto human cloth in the digital space. They test the garment in the physical world with motions and justify their result in both digital space and physical space.

\subsection{Mesh Adversarial Attacks}
A 2D object can be considered as a projection of its 3D model. Therefore, attacking from 3D space and then mapping to 2D space can be seen as a way of augmenting perturbation space. 
In recent years, different adversarial attack schemes for 3D meshes have been proposed. For instance, Tsai et al. \cite{tsai2020robust} perturbs the position of point clouds to generate an adversarial mesh that fools 3D shape classifiers. Ti et al. \cite{liu2018beyond} generate adversarial attacks by modeling the pixels in natural images as an interaction result of lighting condition and the physical scene, such that the pixels can maintain their natural appearance. More recently, Xiao et al. \cite{xiao2019meshadv} and Zeng et al. \cite{zeng2019adversarial} generate adversarial samples by altering the physical parameters ($e.g. $ illumination) of rendering results from target objects. They generate meshes with negligible perturbations to the texture and show that under certain rendering assumptions ($e.g.$ fixed camera view), the adversarial mesh can deceive state-of-the-art classifiers and detectors. Overall, most existing works perturb an image's global texture, while the idea of generating an adversarial sub-region/patch remains unexplored in the 3D mesh domain.

\section{The Proposed Framework}


\begin{figure*}[t]
\centering
\includegraphics[width=1\linewidth]{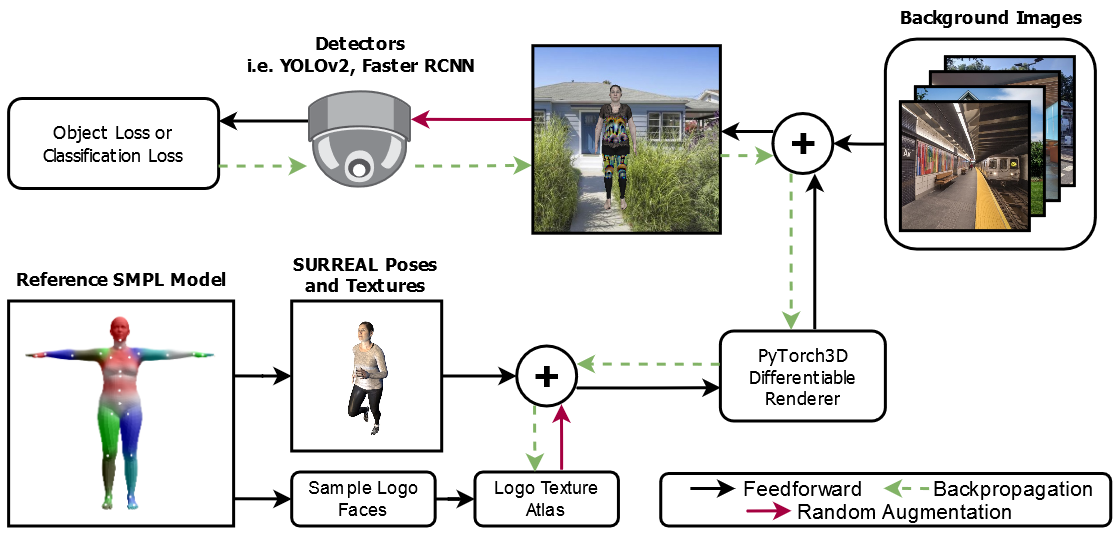}
\caption{The 3D adversarial logo pipeline. We start with the reference SMPL \cite{SMPL:2015} model, and sample its faces to form a desired logo shape. The SURREAL \cite{varol17_surreal} dataset is used to create a wide variety of body poses and mesh textures during training and testing. The logo texture atlas defined by the sampled faces is then randomly perturbed, and appended to our human meshes. These meshes are rendered using PyTorch3D, and imposed upon real-world background images. Finally, the synthesized images are fed through various object detectors, which allows for the computation of disappearance loss (\ref{sec:loss}). As the whole pipeline is differentiable, we backpropagate from the losses, to the ``Logo Texture Atlas'' along the green arrows.
}
\label{fig:Training}
\end{figure*}

In this section, we seek a concrete solution to the 3D adversarial logo attack, with the following goals in mind:

\begin{itemize}

\item [$\bullet$] The 3D adversarial logo is \textit{universal}: for every distinct human mesh, we will apply the logo in a manner such that there is little discrepancy between logos on different meshes. Our use of the SMPL model will facilitate universality among the applied logos.

\item [$\bullet$] The adversarial training is \textit{differentiable}: we will modify the logo's texture atlas via end-to-end loss backpropagation. The major challenge is to replace a traditional discrete renderer with a differentiable one.
    
\item [$\bullet$] The trained 3D adversarial logo is \textit{robust}: to fully exploit our 3D pipeline, we will create an augmented training procedure that utilizes many camera angles, body poses, background images, and random image perturbations. We hope the resulting adversarial logo will be robust in real-world scenarios, unlike 2D patch attacks.
\end{itemize}

Our 3D adversarial logo attack pipeline is outlined in Figure \ref{fig:Training}. In the training procedure, we first sample faces on the reference human mesh to construct the desired logo shape.  Even though in the texture atlas representation, each face can be represented by an $R \times R$ texture map, and the texture value at particular points can be evaluated using barycentric interpolation, in our case, however, we use a resolution of $R=1$ for each face. This setting falls into a piecewise constant function of colors defined over each face in the mesh. We found that the interpolation step for higher resolutions caused our gradient to be weakened for meshes with many faces. We apply random perturbations (brightness, contrast, noise) to the logo's texture atlas, then attach the logo to each human mesh. The meshes are then rendered using PyTorch3D, and imposed onto real-world background images. Finally, the synthesized images are streamed through object detectors for adversarial training.



Due to end-to-end differentiability, the training process updates the 3D adversarial logo texture atlas via backpropagation. Within one epoch, the above process will be conducted on all training meshes and background images.

\subsection{Mesh Acquisition via SMPL Body Model}
To alleviate the problem of overfitting to certain meshes and to enrich our dataset, we use the SMPL body model \cite{SMPL:2015} to generate human meshes. The SMPL model is a kind of parametric 3D body model that is learned from thousands of 3D body scans. There are 10 parameters to control the human body shapes, and 72 parameters to control the locations and orientations of the 24 major human joints. These 82 parameters can be acquired in datasets like the SURREAL Dataset \cite{varol17_surreal}, which contains a large number of different shapes and pose parameters for the SMPL model. We can generate infinitely many human meshes with different poses, texture mappings, and body shapes using the SMPL model. Another advantage of these human meshes is their topological consistency. The 3D adversarial patch, when trained using human meshes generated by SMPL model, only need to be constructed once, and the corresponding topology can be assigned to every mesh simply with SMPL model generation. This advantage enables us to conduct fair analysis on our adversarial attack model’s performance over different meshes.

\subsection{Differentiable Rendering}
\label{sec:differentiable:renderer}
A differentiable renderer can take meshes and texture maps as input, and produce a 2D rendered image using differentiable operations. This allows gradients of 3D meshes and their texture maps to propagate through their corresponding 2D image projections. Differentiable rendering has been used in many 3D optimization tasks, such as pose estimation \cite{yuanlu2019pose, pavlakos2018pose}, object reconstruction \cite{wenzheng2019reconstruction, shubham2017reconstruction}, and texture fitting \cite{kato2018neural}. Our work is built upon a specific renderer called PyTorch3D \cite{ravi2020pytorch3d}, which is implemented using PyTorch \cite{pytorch}. PyTorch3D allows us to conveniently represent our 3D adversarial logo as a texture atlas, which will be optimized during backpropagation. 


\subsection{Adversarial Loss Functions}
\label{sec:loss}

The aim of our work is to generate a 3D adversarial logo that, when applied to a human mesh, can fool the object detector when it is rendered into a 2D image. We will now discuss the loss functions employed to achieve this goal.

\paragraph{Disappearance Loss}
To fool an object detector is to diminish the confidence within bounding boxes that contain the target object. We exploit the disappearance loss \cite{eykholt2018physical}, which takes the maximum confidence of all bounding boxes that contain the target object:
\begin{equation}    
    \mathrm{DIS}(\mathcal{I}, y) = \max_{b \in B} \mathrm{Conf}(\mathcal{O_\theta}(\mathcal{I}), b, y),
    \label{eq:loss:DIS}
\end{equation}
where $\mathrm{Conf}(\cdot)$ computes the confidence that a bounding box prediction $b$, given by object detector $\mathcal{O}_\theta$, corresponds to class label $y$. The object detector operates on an input image $\mathcal{I}$. In our case, we hope to minimize the maximum confidence of human detections from $\mathcal{O}_\theta$.


\paragraph{Total Variance Loss}
Patch-based adversarial attacks are substantially weaker in the real-world when the resulting patch contains high variance among neighboring pixels. In order to increase our attack robustness, we apply a smoothing loss to the 3D adversarial logo. In previous works involving 2D patches, pixel-wise total variation loss is enforced \cite{sharif2016accessorize, eykholt2018physical}
\begin{equation}
    \mathrm{TV}(r) = \sum_{i, j}|r_{i+1,j} - r_{i,j}| + |r_{i,j+1} - r_{i,j}|
\end{equation}
where $r_{i,j}$ is the pixel value at coordinate $(i, j)$ in a 2D image $r$. However, in our case the patch is not defined in the conventional 2D image representation, but rather as a texture atlas. We apply a mesh-based total variation loss described in \cite{huayan2015meshtv}, which is only suitable for piecewise constant functions. Our logo's texture atlas has resolution $1$; hence, it is defined over a piecewise constant function $C$ per face. Given triangular face $\Delta$, let $C(\Delta$) indicate the three dimensional color vector for that particular face.
The total variation loss can now be formulated as
\begin{equation}
\label{eq:TV_loss}
    \mathrm{TV}({\mathcal{L}}) = \sum_{e \in \mathcal{L}'} |e| \cdot |C(\Delta_1) - C(\Delta_2)|
\end{equation}
where $\mathcal{L}'$ is the collection of non-boundary edges in the 3D adversarial logo, and $\Delta_1$, $\Delta_2$ are the triangular faces conjoined along their common edge $e$ with length $|e|$.


The overall training loss we are minimizing is composed of the above two losses ($\lambda_{\mathrm{DIS}}$ and $\lambda_{\mathrm{TV}}$ are hyperparameters):
\begin{equation}
  \mathcal{L}_{\mathrm{adv}} = \lambda_{\mathrm{DIS}} \mathrm{DIS}(\mathcal{I}, y) + \lambda_{\mathrm{TV}}\mathrm{TV}({\mathcal{L}}) 
  \label{eq:loss:overall}
\end{equation}

\section{Experiments and Results}


\subsection{Dataset Preparation}




\paragraph{Background Images}
In the interest of synthesizing realistic renderings, we sample background images from the MIT Places Database \cite{zhou2017places}. We selected images across a diverse set of indoor and outdoor categories, such as \textit{beach, bedroom, boardwalk, courthouse, driveway, house, kitchen}, and more. A total of $1,400$ training and $1,200$ testing backgrounds were collected. During both training and testing, we render human meshes at varying viewing angles and spatial locations. For most of our experiments (apart from single-angle training, see Section \ref{sec:single:multiple}), we sample 5 viewing angles, which effectively scales our training set to a size of 7,000 images. We demonstrate that these images are sufficient for robust adversarial patch training in Section \ref{sec:results_and_analysis}.

\begin{figure*}[ht]
\centering
\includegraphics[width=0.95\linewidth]{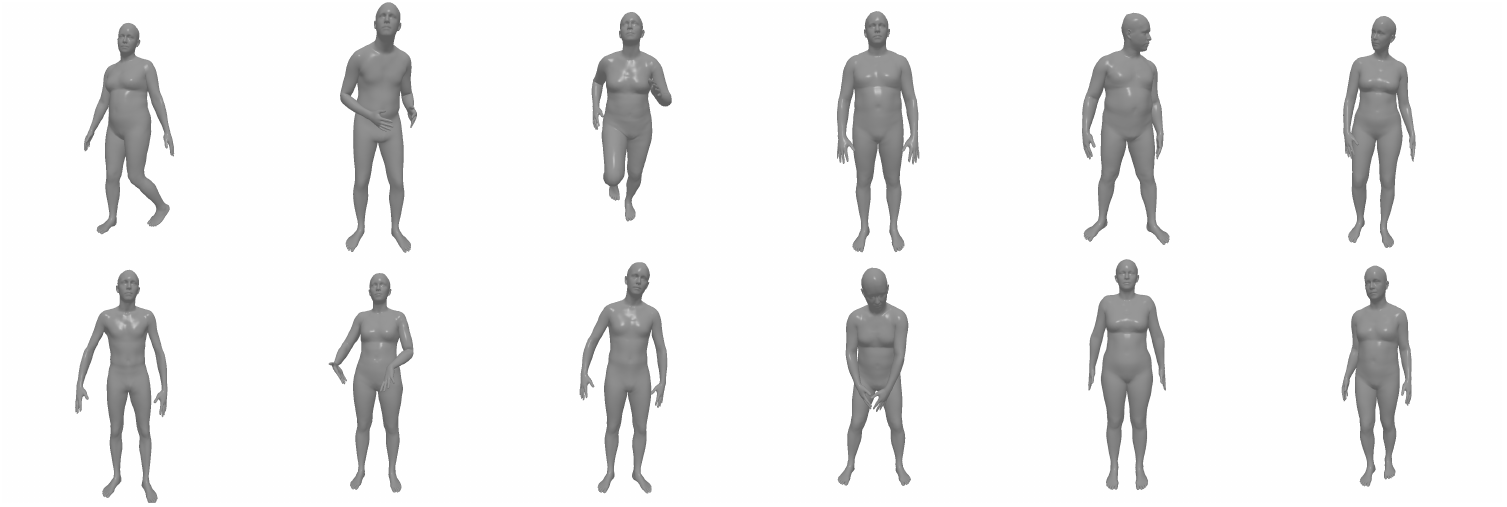}
\caption{A sample of the 3D human meshes that we use to train our adversarial logos. Our meshes are defined using the SMPL human body model, and the poses are sampled from the SURREAL dataset. 
\label{fig:poses}}
\end{figure*}

\paragraph{Human Meshes and Texture Maps} 
We sampled twelve 3D human meshes and texture maps from the SURREAL dataset \cite{varol17_surreal}. As the meshes are all created using the SMPL model, we are guaranteed topological consistency between each mesh. Consequently, our logo's texture atlas can be directly applied to all SMPL meshes without the need for correspondence mapping. The human meshes that we sampled display varying poses, body shapes, and surface deformations. Our human meshes are shown in Figure \ref{fig:poses}.
 
We found that our adversarial logo was unable to express enough detail under the resolution of 6,890 vertices in the SMPL model. In order to train an intricate adversarial logo, we apply a preliminary subdivision step to the meshes in our dataset. The Subdivision Surface Modifier routine in \textit{Blender} \cite{blender2020} was used with parameters ``simple'' and ``levels'' equal to 1. The resulting meshes remain topologically consistent and now contain 27,578 vertices. We found that further subdivision was unnecessary, as it greatly increases the training time, while only providing slightly more detail.

The ``Sample Logo Faces'' step in our pipeline (\ref{fig:Training}) involves manual sampling of triangular faces in Blender. We export a list of face indices that delineate the region we wish to perturb in the mesh texture atlas. This list of faces is universal among all human meshes that are derived from the SMPL model. We manually select regions over meshes to be attacked according to the heatmaps of detection models, which usually concentrate on chests and thighs.

\begin{figure*}[t]
\centering
\includegraphics[width=1\linewidth]{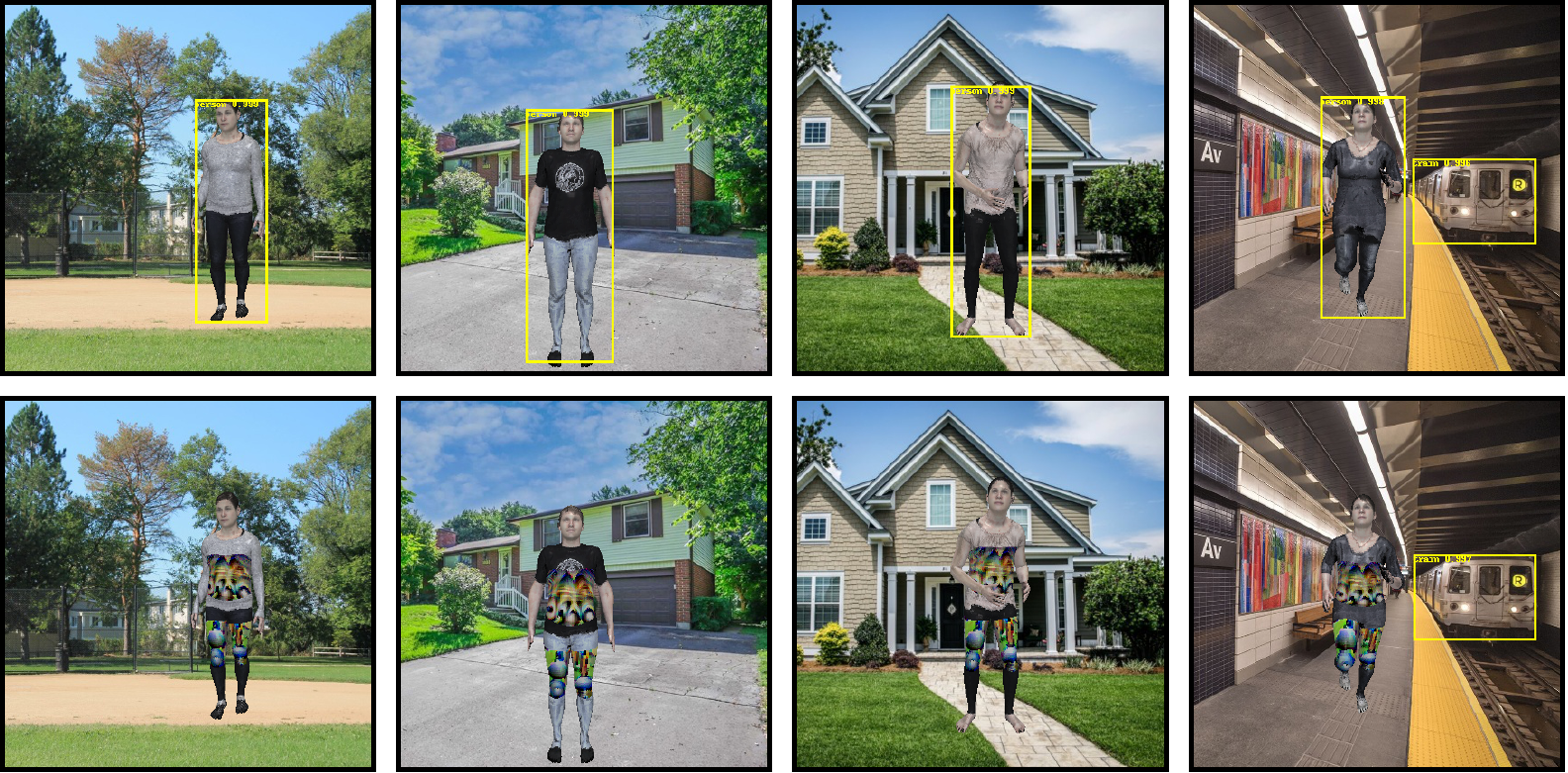}
\caption{Examples of our adversarial attack against Faster R-CNN. The first row contains human meshes without any adversarial perturbation; Faster R-CNN is $99\%$ confident in its human predictions in these images. The second row displays the cloaking effect of an adversarial patch trained by the pipeline outlined in Figure \ref{fig:Training}. To bolster our attack robustness, we train and test our adversarial logos on meshes with a diverse set of surface-level and body-level deformations. The figure above features running, walking, and idle poses on meshes of various shapes and sizes, which are sampled from the SURREAL dataset. We even observe attack success for partially occluded adversarial textures ($e.g.$ the third column).
\label{fig:res}}
\end{figure*}

 
\subsection{Implementation Details}
\label{sec:experiments:settings}
All experiments are implemented in PyTorch 1.6.0, along with PyTorch3D 0.2.5. The scene parameters include: camera distance (2.2), elevation (6.0), rasterization blur radius (0.0), image size (416x416), and one point light with default color. For data augmentation, we apply random brightness and contrast uniformly generated from 0 to 1, and noise uniformly generated from $-0.1$ to $0.1$. All three are added pixel-wise to the rendered images. Additionally, the rendered meshes are randomly translated by $-50$ to $50$ pixels along the height and width axes in the background images. The training is conducted on one Nvidia GTX\,1080TI GPU. We use the SGD optimizer with an initial learning rate of 0.1, which is decayed by a factor of 0.1 every 10 epochs.


During all experiments, weight parameters are set to be $\lambda_{\mathrm{DIS}}=1.0$ and $\lambda_{\mathrm{TV}}=2.5$ in \eqref{eq:loss:overall} unless otherwise specified. When training against YoloV2, the batch size is $16$ during single-angle training and $8$ during multi-angle training. For training against Faster R-CNN we use a batch size of $1$. The nature of our single-angle and multi-angle training experiments are outlined in Section \ref{sec:single:multiple}. The number of epochs used is $100$ unless otherwise specified. The default object detectors used are YOLOv2 \cite{redmon2017yolo9000} and Faster R-CNN \cite{shaoqing2015fasterRCNN}, with confidence thresholds set to 0.6 for both detectors. 
 
\subsection{Experiment Results and Analysis}
\label{sec:results_and_analysis}
\subsubsection{Training Schemes}
\label{sec:single:multiple}
\begin{figure}[h]
\centering
\includegraphics[width=1\linewidth]{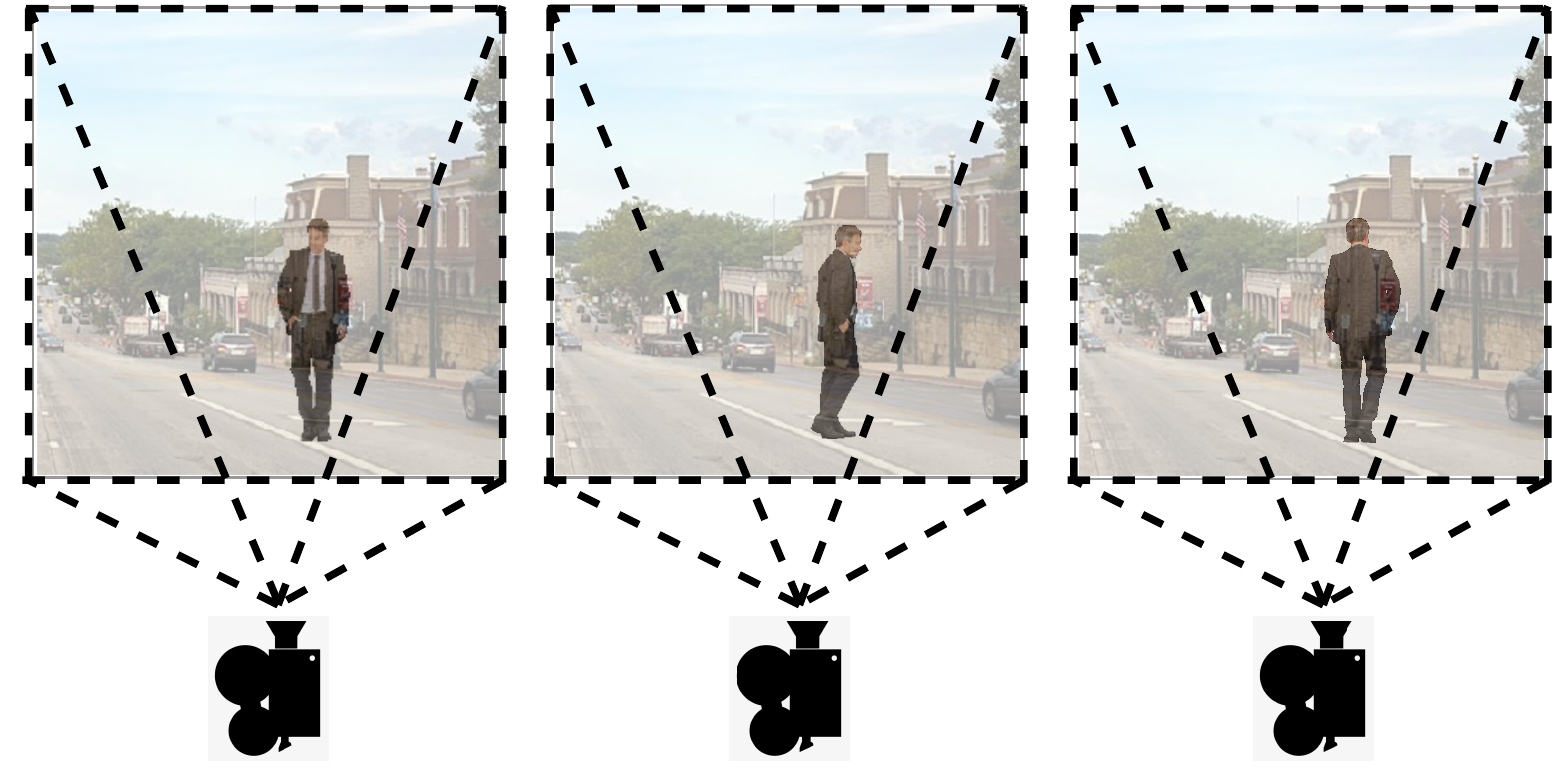}
\caption{Examples of camera angle settings. From left to right: $0$ degrees, $-90$ degrees and $+90$ degrees for one background image and one human model.}
\label{fig:view}
\end{figure}

\paragraph{Single-angle training} We first apply our 3D adversarial attack pipeline to images rendered at a single angle. More specifically, the camera's azimuth angle relative to the human meshes is $0$ degrees. We synthesize 2D images by imposing the rendered meshes onto our collection of background images. The synthesized images used during testing follow the same scheme, but with a separate set of test backgrounds and human meshes. We will refer to these as \textit{unseen} backgrounds and meshes. The attack success rate denotes the ratio of testing samples where the target detector failed to detect the rendered human mesh. A visualization of the single-angle renderings can be seen in Figure \ref{fig:patches}.


\paragraph{Multi-angle training} In the interest of real-world attack robustness, we extend our pipeline to perform joint multi-angle training. We render the human meshes from azimuth angles -10, -5, 0, 5, and 10 during training; however, during testing we use all 21 integer angles in $[-10, 10]$. Under this setting, the training set and testing set are enlarged by a factor of $5$ and $21$ respectively. We compute our multi-angle success rate by averaging the success rates across all $21$ views. Results are summarized in the last column of Table \ref{table:angle}. As can be seen in Table \ref{table:angle}, a lower success rate implies that the multi-angle attack is more challenging compared to the single-angle attack. Examples of human meshes rendered at various camera angles can be seen in Figure \ref{fig:extreme_angles}. Note that, when rotating the camera, the background image remains static.

The numbers we report in Table \ref{table:angle} are consistent with our visual results. A sample of the images from our multi-angle training are shown in Figure \ref{fig:res}. 
As one can observe, our adversarial patches can mislead the pre-trained object detectors and make our human meshes unrecognizable.


\begin{table}[ht]
\begin{center}
\caption{Results for various patches in single-angle and multi-angle training. Baseline attack rates are denoted as the ``None'' patch.}
\label{table:angle}
\begin{threeparttable}
\resizebox{0.48\textwidth}{!}{
\begin{tabular}{c|c|c|c}
\hline
\multirow{2}{*}{Object Detector} & \multirow{2}{*}{Patch} & \multicolumn{2}{c}{\textbf{Attack Success Rate}} \\ \cline{3-4} 
& & Single-angle & Multi-angle \\
\hline
YoloV2 & None & 0.01 & 0.01  \\
YoloV2 & Letter G & 0.98 & 0.86 \\
YoloV2 & Smiley Face & 0.98 & 0.88 \\
YoloV2 & Chest + Thighs & 0.99 & 0.93 \\
\hline
Faster R-CNN & None & 0.01 & 0.01 \\
Faster R-CNN & Letter G & 0.68 & 0.62 \\
Faster R-CNN & Smiley Face & 0.51 & 0.47 \\
Faster R-CNN & Chest + Thighs & 0.99 & 0.91 \\
\hline
\end{tabular}}
\end{threeparttable}
\end{center}
\end{table}



\subsubsection{Attacking unseen camera angles}

\begin{figure}[ht]
\centering
\includegraphics[width=\linewidth]{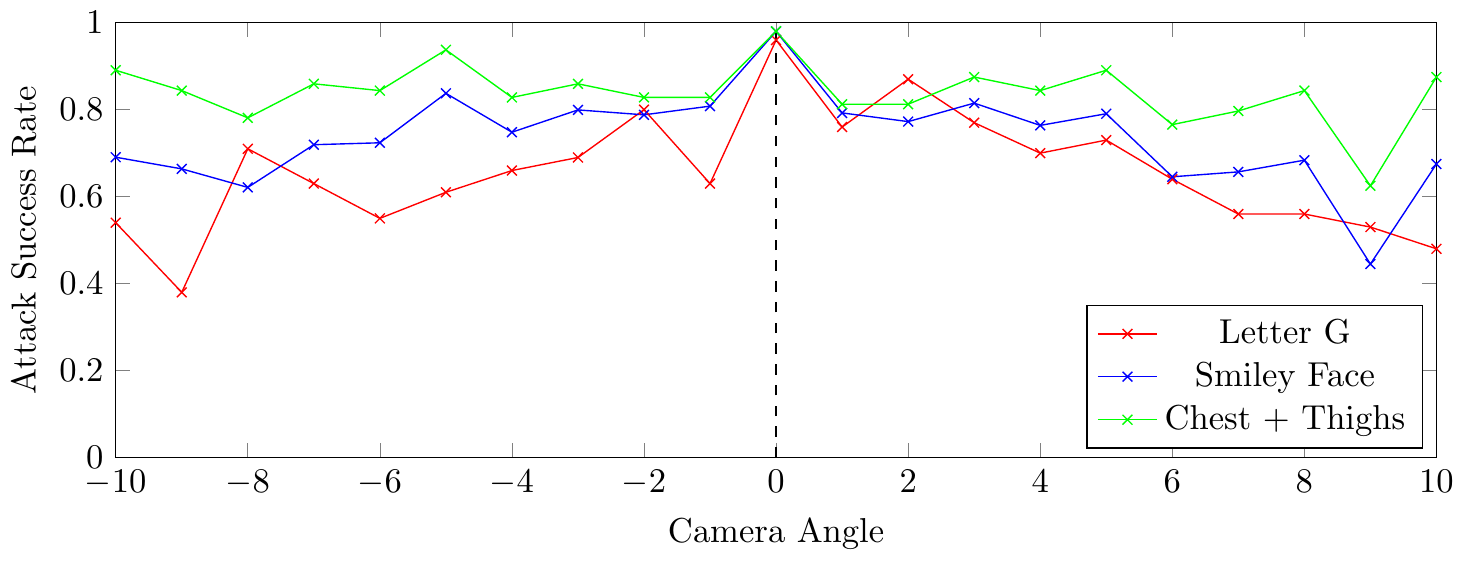}
\caption{The attack success rate for various adversarial patches against YoloV2. The patches are trained on a single viewing angle (0 degrees), and tested against 21 viewing angles.}
\label{fig:yolov2_single_angle}
\end{figure}

\paragraph{Single-angle training against unseen camera angles}
To prove our method is robust against 3D rotations, we conduct a multi-angle attack with single-angle training. We first train at $0$ degrees, but use $21$ angles in $[-10,10]$ to attack the detectors. Results shown in Figure \ref{fig:yolov2_single_angle} show that our method is stable against small camera angle perturbations. Figure \ref{fig:intro} provides an example where our 3D adversarial logo hides a human mesh from the detectors. Nevertheless, our method is not affected by minor pixel-level changes that could collapse 2D patch-based attacks.

\paragraph{Multi-angle training against unseen camera angles} 

\begin{figure}[ht]
\centering
\includegraphics[width=1\linewidth]{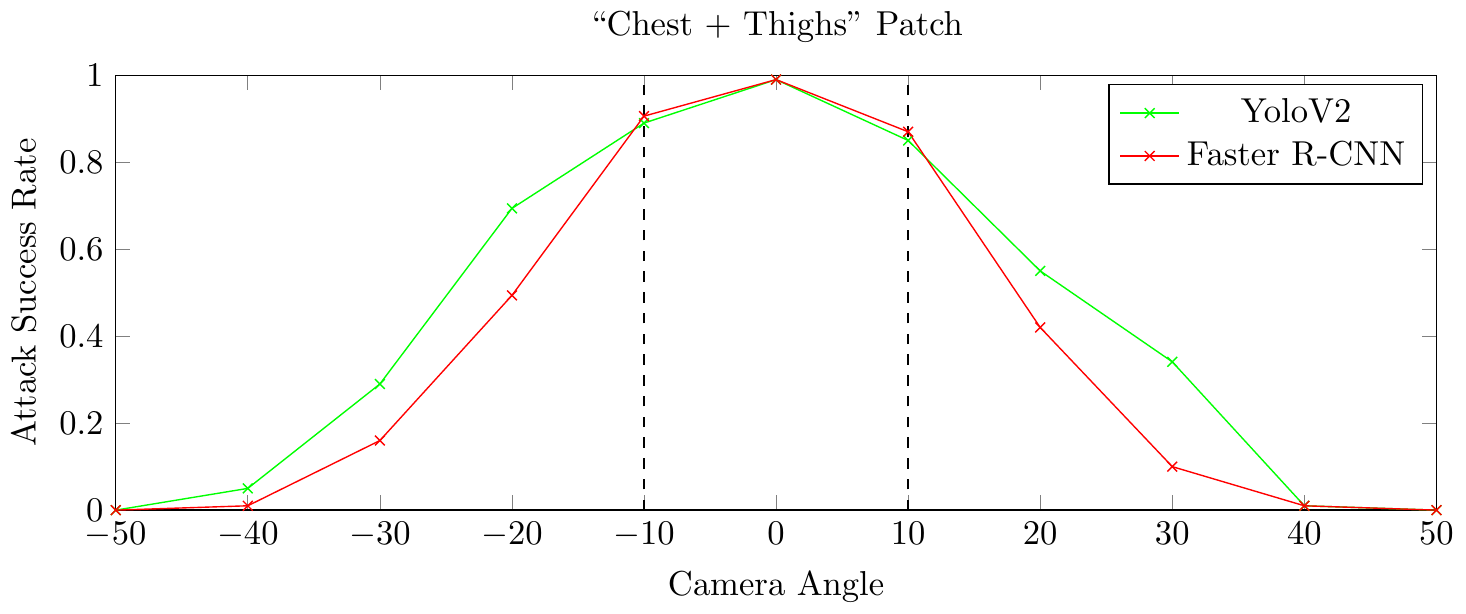}
\caption{The attack success rate against YoloV2 and Faster R-CNN for wide viewing angles. The ``Chest + Thighs'' patch was trained under 5 camera views in $[-10,10]$ as highlighted by the dotted lines. There exists a massive performance drop when the viewing angle is relatively large.}
\label{fig:multi-wider}
\end{figure}

\begin{figure}[ht]
\centering
\includegraphics[width=1\linewidth]{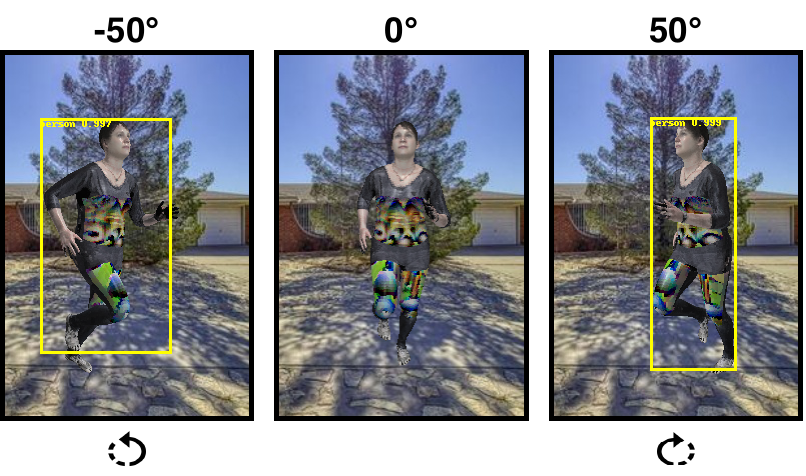}
\caption{An example of the limitation of our adversarial patches. Under extreme viewing angles or occlusions, our patch loses attacking robustness.}
\label{fig:extreme_angles}
\end{figure}

\begin{figure}[ht]
\centering
\includegraphics[width=0.99\linewidth]{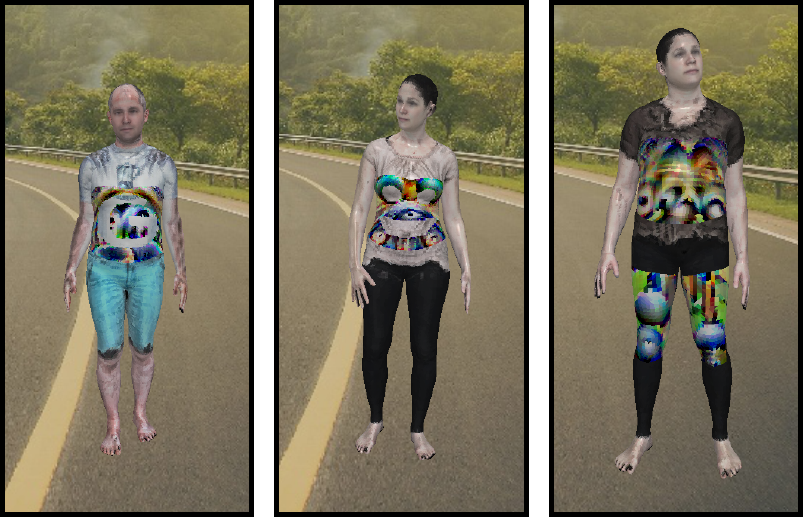}
\caption{Patches ``Letter G,'' ``Smiley Face,'' and ``Chest + Thighs'' trained against Faster R-CNN and applied to various 3D human meshes. The meshes are rendered with an azimuth angle of 0, then imposed on a highway background image. The three patches are defined by 2,426, 2,427, and 4,691 faces respectively.}
\label{fig:patches}
\end{figure}

We extend our experiments to test robustness under more camera angles. After training with $5$ angles in $[-10,10]$ degrees, we attack the detectors using angle views in $[-50,50]$ degrees with an increment of 10. Figure \ref{fig:multi-wider} is plotted based on our attack success rate over all test images on both YoloV2 and Faster R-CNN. The plot in Figure \ref{fig:multi-wider} reveals the limitations of our adversarial patches. We observe a decaying curve that converges to a success rate of $0\%$. This is expected because the patch becomes less visible as the camera angle deviates from $0$ degrees.

\vspace{-2mm}

\vspace{-2mm}

\subsubsection{Shape adaptivity}
While our attacking pipeline is not restricted to a particular patch shape, the results from different patches reveals that shape and size are non-negligible factors in the attack success rate. As seen in Table \ref{table:angle} and Figure \ref{fig:yolov2_single_angle}, there is a significant contrast in attacking performance between the various patches. When attacking Faster R-CNN in particular, we observe the necessity for a a larger patch ($e.g.$ Chest + Thighs). In Figure \ref{fig:patches}, the relative sizes of the various patches can be seen.

\subsubsection{Blackbox transferability}
\begin{figure}[ht]
\centering
\includegraphics[width=1\linewidth]{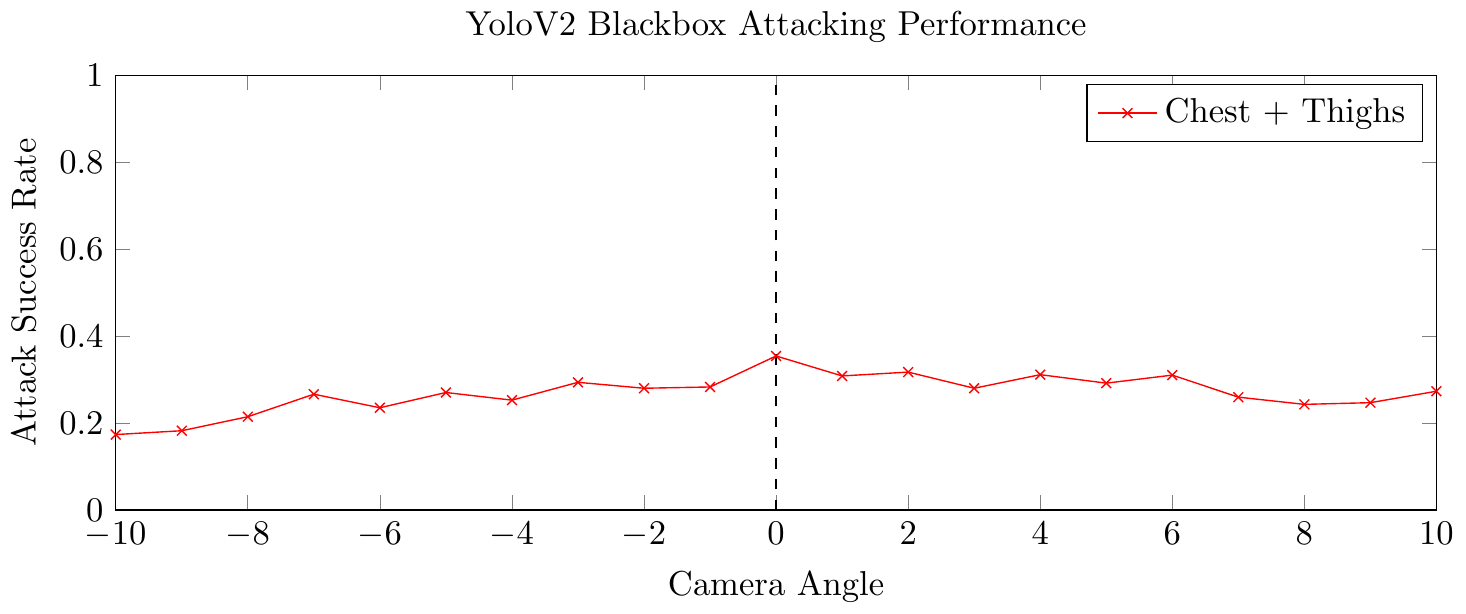}
\caption{The attack success rate against YoloV2 of a ``Chest + Thighs'' patch trained on Faster R-CNN. One could not observe a relatively high performance, yet the attack success rate is robust to every angles.}
\label{fig:yolo_blackbox}
\vspace{-3mm}
\end{figure}

To test the generalizability of our adversarial patches, we choose Faster R-CNN as our whitebox during training, and YoloV2 as our \textit{unseen} detector to perform blackbox attacking. We generate the ``Chest + Thighs'' patch under the multi-angle training scheme mentioned in (\ref{sec:single:multiple}). Then, we attempt to fool YoloV2 with this patch. In Figure \ref{fig:yolo_blackbox}, the transferred attack success rate can be seen for all angles in $[-10, 10]$. Despite not being specifically optimized for YoloV2, our patch is able to fool the detector in many cases.


\subsubsection{Ablation study of total variation loss}

\begin{figure}[ht]
\centering
\includegraphics[width=1\linewidth]{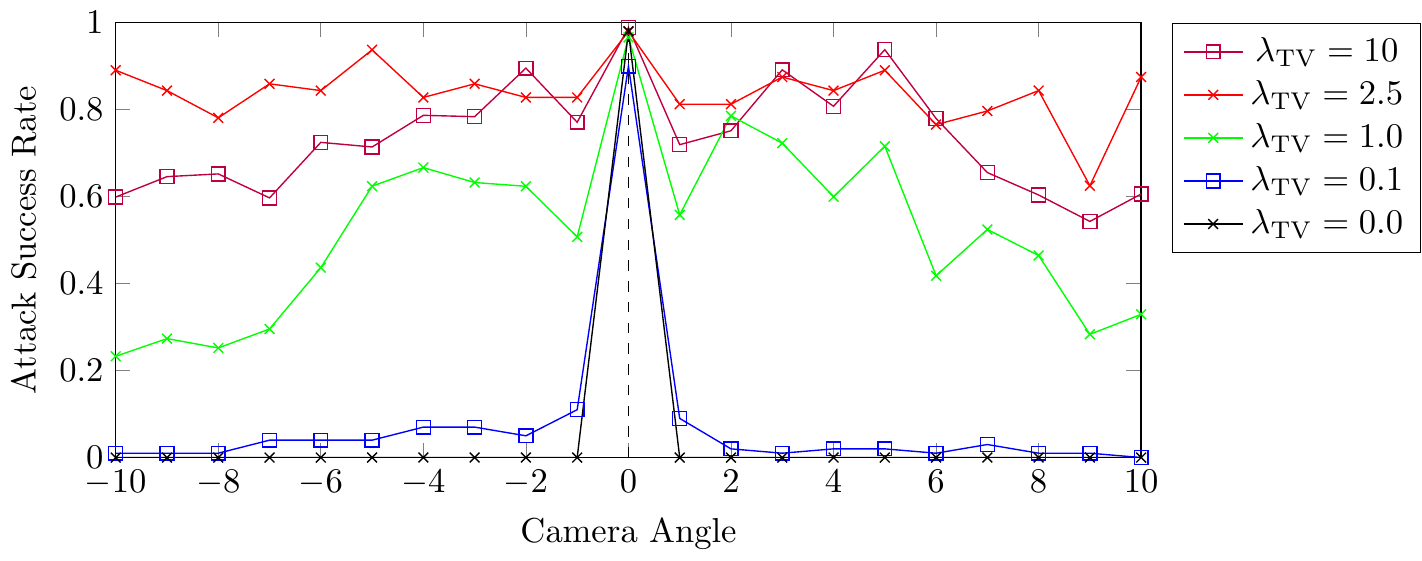}
\caption{The performance of various $\lambda_{\mathrm{TV}}$ values on a ``Chest + Thighs'' patch when trained against YoloV2. We trained each patch under the identical setting except for the setting of $\lambda_{TV}$. The plot is generated from multi-angle testing ($[-10,10]$ degrees) with single-angle training (0 degrees) on one human mesh and one unseen human mesh.  }
\label{fig:TV_loss}
\vspace{-3mm}
\end{figure}

Since our 3D Adversarial Logo is not defined in 2D space, we change the formulation of TV loss into (\ref{eq:TV_loss}) as a smoothness constraint. To address the necessity of smoothing the adversarial patches, we performed our attack under different weights of total variation loss (\ref{eq:TV_loss}) by changing $\lambda_{TV}$, including $\lambda_{TV}=0$. The results in Figure \ref{fig:TV_loss} show that the total variation affects our attack success rate significantly. The adversarial patch generated without this smoothing penalty ($\lambda_{TV}=0$) is entirely unable to attack unseen camera angles. This is due to the extreme amount of fine detail present in a patch with high total variation. Moreover, we found $\lambda_{TV}=2.5$ yields the maximum attack success rate under our setting and thereby we apply the weight to most of our experiments aforementioned.

\vfill

\section{Conclusion}

We have presented our novel 3D adversarial logo attack on human meshes. A logo shape sampled from a reference human mesh is used to generate an adversarial texture atlas, which is transferable to a variety of human meshes from the SMPL model. Due to differentiable rendering, the update back to the logo texture atlas is shape-free, mesh-free, and angle-free, leading to a stable attack success rate under different angle views with different human models and logo shapes. We comprehensively show our attacking performance under two different whitebox attacking scenarios and justify our success. Our method enables one to create diverse adversarial patches that are more robust in the physical world. In future work, We hope to explore the printability of our adversarial texture atlas, and its performance in the realistic physical world when worn by humans. We would also like to explore joint optimization of both the texture atlas and the human poses that consistently fool the object detector. Right now our attack only operates on static poses, it is an open question on how to robustly attack humans in a video with drastic pose changes. Our work has the potential to extend to versatile adversarial attack scenarios. It is possible to transfer our attack to unseen 3D human models that are not from the SMPL model. 


\vspace{1em}

\noindent {\it\bfseries Acknowledgements}: This research was supported in part by a grant from NIH - R01GM117594, in part from the Peter O’Donnell Foundation, and in part from a grant from the Army Research Office accomplished under Cooperative Agreement Number W911NF-19-2-0333. The views and conclusions contained in this document are those of the authors and should not be interpreted as representing the official policies, either expressed or implied, of the Army Research Office or the U.S. Government. The U.S. Government is authorized to reproduce and distribute reprints for Government purposes notwithstanding any copyright notation herein.

{\small
\bibliographystyle{ieee_fullname}
\bibliography{ref}

\begin{thebibliography}{10}\itemsep=-1pt

\bibitem{azulay2018deep}
Aharon Azulay and Yair Weiss.
\newblock Why do deep convolutional networks generalize so poorly to small
  image transformations?
\newblock {\em arXiv preprint arXiv:1805.12177}, 2018.

\bibitem{blender2020}
{Blender Online Community}.
\newblock {\em Blender - a 3D modelling and rendering package}.
\newblock Blender Foundation, Blender Institute, Amsterdam, 2020.

\bibitem{brown2017adversarial}
Tom~B Brown, Dandelion Man{\'e}, Aurko Roy, Mart{\'\i}n Abadi, and Justin
  Gilmer.
\newblock Adversarial patch.
\newblock {\em arXiv preprint arXiv:1712.09665}, 2017.

\bibitem{chen2018shapeshifter}
Shang-Tse Chen, Cory Cornelius, Jason Martin, and Duen Horng~Polo Chau.
\newblock Shapeshifter: Robust physical adversarial attack on faster r-cnn
  object detector.
\newblock In {\em Joint European Conference on Machine Learning and Knowledge
  Discovery in Databases}, pages 52--68. Springer, 2018.

\bibitem{Chen_2020_CVPR}
Tianlong Chen, Sijia Liu, Shiyu Chang, Yu Cheng, Lisa Amini, and Zhangyang
  Wang.
\newblock Adversarial robustness: From self-supervised pre-training to
  fine-tuning.
\newblock In {\em The IEEE/CVF Conference on Computer Vision and Pattern
  Recognition (CVPR)}, June 2020.

\bibitem{wenzheng2019reconstruction}
Wenzheng Chen, Jun Gao, Huan Ling, Edward~J. Smith, Jaakko Lehtinen, Alec
  Jacobson, and Sanja Fidler.
\newblock Learning to predict 3d objects with an interpolation-based
  differentiable renderer.
\newblock {\em CoRR}, abs/1908.01210, 2019.

\bibitem{engstrom2019exploring}
Logan Engstrom, Brandon Tran, Dimitris Tsipras, Ludwig Schmidt, and Aleksander
  Madry.
\newblock Exploring the landscape of spatial robustness.
\newblock In {\em International Conference on Machine Learning}, pages
  1802--1811, 2019.

\bibitem{eykholt2018physical}
Kevin Eykholt, Ivan Evtimov, Earlence Fernandes, Bo Li, Amir Rahmati, Florian
  Tramer, Atul Prakash, Tadayoshi Kohno, and Dawn Song.
\newblock Physical adversarial examples for object detectors.
\newblock {\em arXiv preprint arXiv:1807.07769}, 2018.

\bibitem{girshick2015fast}
Ross Girshick.
\newblock Fast r-cnn.
\newblock In {\em Proceedings of the IEEE international conference on computer
  vision}, pages 1440--1448, 2015.

\bibitem{goodfellow2014explaining}
Ian Goodfellow, Jonathon Shlens, and Christian Szegedy.
\newblock Explaining and harnessing adversarial examples.
\newblock In {\em International Conference on Learning Representations (ICLR)},
  2015.

\bibitem{gui2019ATMC}
Shupeng Gui, Haotao Wang, Haichuan Yang, Chen Yu, Zhangyang Wang, and Ji Liu.
\newblock Model compression with adversarial robustness: A unified optimization
  framework.
\newblock In {\em Proceedings of the 33rd Conference on Neural Information
  Processing Systems}, 2019.

\bibitem{tkhu2019triplewins}
Ting{-}Kuei Hu, Tianlong Chen, Haotao Wang, and Zhangyang Wang.
\newblock Triple wins: Boosting accuracy, robustness and efficiency together by
  enabling input-adaptive inference.
\newblock In {\em ICLR}, 2020.

\bibitem{huang2020universal}
Lifeng Huang, Chengying Gao, Yuyin Zhou, Cihang Xie, Alan~L Yuille, Changqing
  Zou, and Ning Liu.
\newblock Universal physical camouflage attacks on object detectors.
\newblock In {\em Proceedings of the IEEE/CVF Conference on Computer Vision and
  Pattern Recognition}, pages 720--729, 2020.

\bibitem{kato2018neural}
Hiroharu Kato, Yoshitaka Ushiku, and Tatsuya Harada.
\newblock Neural 3d mesh renderer.
\newblock In {\em Proceedings of the IEEE Conference on Computer Vision and
  Pattern Recognition}, pages 3907--3916, 2018.

\bibitem{liu2018beyond}
Hsueh-Ti~Derek Liu, Michael Tao, Chun-Liang Li, Derek Nowrouzezahrai, and Alec
  Jacobson.
\newblock Beyond pixel norm-balls: Parametric adversaries using an analytically
  differentiable renderer.
\newblock {\em arXiv preprint arXiv:1808.02651}, 2018.

\bibitem{SMPL:2015}
Matthew Loper, Naureen Mahmood, Javier Romero, Gerard Pons-Moll, and Michael~J.
  Black.
\newblock {SMPL}: A skinned multi-person linear model.
\newblock {\em ACM Trans. Graphics (Proc. SIGGRAPH Asia)}, 34(6):248:1--248:16,
  Oct. 2015.

\bibitem{lu2017no}
Jiajun Lu, Hussein Sibai, Evan Fabry, and David Forsyth.
\newblock No need to worry about adversarial examples in object detection in
  autonomous vehicles.
\newblock {\em arXiv preprint arXiv:1707.03501}, 2017.

\bibitem{nguyen2018rendernet}
Thu~H Nguyen-Phuoc, Chuan Li, Stephen Balaban, and Yongliang Yang.
\newblock Rendernet: A deep convolutional network for differentiable rendering
  from 3d shapes.
\newblock In {\em Advances in Neural Information Processing Systems}, pages
  7891--7901, 2018.

\bibitem{pytorch}
Adam Paszke, Sam Gross, Francisco Massa, Adam Lerer, James Bradbury, Gregory
  Chanan, Trevor Killeen, Zeming Lin, Natalia Gimelshein, Luca Antiga, Alban
  Desmaison, Andreas Kopf, Edward Yang, Zachary DeVito, Martin Raison, Alykhan
  Tejani, Sasank Chilamkurthy, Benoit Steiner, Lu Fang, Junjie Bai, and Soumith
  Chintala.
\newblock Pytorch: An imperative style, high-performance deep learning library.
\newblock In H. Wallach, H. Larochelle, A. Beygelzimer, F. d' Alch\'{e}-Buc, E.
  Fox, and R. Garnett, editors, {\em Advances in Neural Information Processing
  Systems 32}, pages 8024--8035. Curran Associates, Inc., 2019.

\bibitem{pavlakos2018pose}
Georgios Pavlakos, Luyang Zhu, Xiaowei Zhou, and Kostas Daniilidis.
\newblock Learning to estimate 3d human pose and shape from a single color
  image.
\newblock {\em CoRR}, abs/1805.04092, 2018.

\bibitem{raj2019learning}
Amit Raj, Cusuh Ham, Connelly Barnes, Vladimir Kim, Jingwan Lu, and James Hays.
\newblock Learning to generate textures on 3d meshes.
\newblock In {\em Proceedings of the IEEE Conference on Computer Vision and
  Pattern Recognition Workshops}, pages 32--38, 2019.

\bibitem{ravi2020pytorch3d}
Nikhila Ravi, Jeremy Reizenstein, David Novotny, Taylor Gordon, Wan-Yen Lo,
  Justin Johnson, and Georgia Gkioxari.
\newblock Accelerating 3d deep learning with pytorch3d.
\newblock {\em arXiv:2007.08501}, 2020.

\bibitem{redmon2017yolo9000}
Joseph Redmon and Ali Farhadi.
\newblock Yolo9000: better, faster, stronger.
\newblock In {\em Proceedings of the IEEE conference on computer vision and
  pattern recognition}, pages 7263--7271, 2017.

\bibitem{shaoqing2015fasterRCNN}
Shaoqing Ren, Kaiming He, Ross Girshick, and Jian Sun.
\newblock Faster r-cnn: Towards real-time object detection with region proposal
  networks.
\newblock In C. Cortes, N.~D. Lawrence, D.~D. Lee, M. Sugiyama, and R. Garnett,
  editors, {\em Advances in Neural Information Processing Systems 28}, pages
  91--99. Curran Associates, Inc., 2015.

\bibitem{sharif2016accessorize}
Mahmood Sharif, Sruti Bhagavatula, Lujo Bauer, and Michael~K Reiter.
\newblock Accessorize to a crime: Real and stealthy attacks on state-of-the-art
  face recognition.
\newblock In {\em Proceedings of the 2016 ACM SIGSAC Conference on Computer and
  Communications Security}, pages 1528--1540. ACM, 2016.

\bibitem{su2015render}
Hao Su, Charles~R Qi, Yangyan Li, and Leonidas~J Guibas.
\newblock Render for cnn: Viewpoint estimation in images using cnns trained
  with rendered 3d model views.
\newblock In {\em Proceedings of the IEEE International Conference on Computer
  Vision}, pages 2686--2694, 2015.

\bibitem{szegedy2013intriguing}
Christian Szegedy, Wojciech Zaremba, Ilya Sutskever, Joan Bruna, Dumitru Erhan,
  Ian Goodfellow, and Rob Fergus.
\newblock Intriguing properties of neural networks.
\newblock In {\em International Conference on Learning Representations (ICLR)},
  2013.

\bibitem{thys2019fooling}
Simen Thys, Wiebe Van~Ranst, and Toon Goedem{\'e}.
\newblock Fooling automated surveillance cameras: adversarial patches to attack
  person detection.
\newblock In {\em Proceedings of the IEEE Conference on Computer Vision and
  Pattern Recognition Workshops}, pages 0--0, 2019.

\bibitem{tramer2017ensemble}
Florian Tram{\`e}r, Alexey Kurakin, Nicolas Papernot, Ian Goodfellow, Dan
  Boneh, and Patrick McDaniel.
\newblock Ensemble adversarial training: Attacks and defenses.
\newblock {\em arXiv preprint arXiv:1705.07204}, 2017.

\bibitem{tsai2020robust}
Tzungyu Tsai, Kaichen Yang, Tsung-Yi Ho, and Yier Jin.
\newblock Robust adversarial objects against deep learning models.
\newblock In {\em Proceedings of the AAAI Conference on Artificial
  Intelligence}, volume~34, pages 954--962, 2020.

\bibitem{shubham2017reconstruction}
Shubham Tulsiani, Tinghui Zhou, Alexei~A. Efros, and Jitendra Malik.
\newblock Multi-view supervision for single-view reconstruction via
  differentiable ray consistency.
\newblock {\em CoRR}, abs/1704.06254, 2017.

\bibitem{varol17_surreal}
G{\"u}l Varol, Javier Romero, Xavier Martin, Naureen Mahmood, Michael~J. Black,
  Ivan Laptev, and Cordelia Schmid.
\newblock Learning from synthetic humans.
\newblock In {\em CVPR}, 2017.

\bibitem{wiyatno2019physical}
Rey~Reza Wiyatno and Anqi Xu.
\newblock Physical adversarial textures that fool visual object tracking.
\newblock In {\em Proceedings of the IEEE International Conference on Computer
  Vision}, pages 4822--4831, 2019.

\bibitem{xiao2019meshadv}
Chaowei Xiao, Dawei Yang, Bo Li, Jia Deng, and Mingyan Liu.
\newblock Meshadv: Adversarial meshes for visual recognition.
\newblock In {\em Proceedings of the IEEE Conference on Computer Vision and
  Pattern Recognition}, pages 6898--6907, 2019.

\bibitem{xu2019evading}
Kaidi Xu, Gaoyuan Zhang, Sijia Liu, Quanfu Fan, Mengshu Sun, Hongge Chen,
  Pin-Yu Chen, Yanzhi Wang, and Xue Lin.
\newblock Evading real-time person detectors by adversarial t-shirt.
\newblock {\em arXiv preprint arXiv:1910.11099}, 2019.

\bibitem{yuanlu2019pose}
Yuanlu Xu, Song{-}Chun Zhu, and Tony Tung.
\newblock Denserac: Joint 3d pose and shape estimation by dense
  render-and-compare.
\newblock {\em CoRR}, abs/1910.00116, 2019.

\bibitem{zeng2019adversarial}
Xiaohui Zeng, Chenxi Liu, Yu-Siang Wang, Weichao Qiu, Lingxi Xie, Yu-Wing Tai,
  Chi-Keung Tang, and Alan~L Yuille.
\newblock Adversarial attacks beyond the image space.
\newblock In {\em Proceedings of the IEEE Conference on Computer Vision and
  Pattern Recognition}, pages 4302--4311, 2019.

\bibitem{huayan2015meshtv}
Huayan Zhang, Chunlin Wu, Juyong Zhang, and Jiansong Deng.
\newblock Variational mesh denoising using total variation and piecewise
  constant function space.
\newblock {\em IEEE Transactions on Visualization and Computer Graphics},
  21:1--1, 07 2015.

\bibitem{zhang2019making}
Richard Zhang.
\newblock Making convolutional networks shift-invariant again.
\newblock In {\em Proceedings of the 36th International Conference on Machine
  Learning}, volume~97 of {\em Proceedings of Machine Learning Research}, pages
  7324--7334, Long Beach, California, USA, 09--15 Jun 2019. PMLR.

\bibitem{zhou2017places}
Bolei Zhou, Agata Lapedriza, Aditya Khosla, Aude Oliva, and Antonio Torralba.
\newblock Places: A 10 million image database for scene recognition.
\newblock {\em IEEE Transactions on Pattern Analysis and Machine Intelligence},
  2017.

\bibitem{zhu2017unpaired}
Jun-Yan Zhu, Taesung Park, Phillip Isola, and Alexei~A Efros.
\newblock Unpaired image-to-image translation using cycle-consistent
  adversarial networks.
\newblock In {\em Proceedings of the IEEE international conference on computer
  vision}, pages 2223--2232, 2017.

\end{thebibliography}
}

\appendix
\newpage
{\centering \Large \textbf{Appendix}}

\section{Camera Distance Study}
\subsection{Attacking unseen camera distances}
\begin{figure}[h]
\centering
\includegraphics[width=1\linewidth]{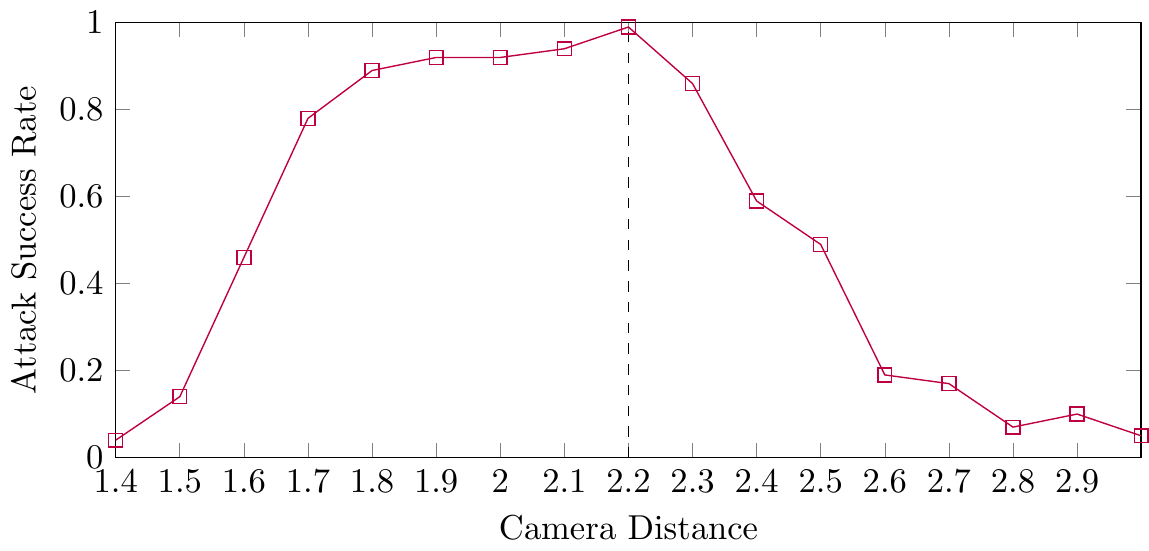}
\caption{The effect of camera distance on attack success rate. A ``Chest + Thighs'' patch was trained against YoloV2 with a camera distance of 2.2, and tested against unseen camera distances in $[1.4, 3.0]$. \label{fig:supp_camera_dist}}
\vspace{-2mm}
\end{figure}

To measure the effect of camera distance on our adversarial attack, we perform two experiments. In the first, we train an adversarial patch using a single camera distance (2.2), and test on unseen camera distances in $[1.4, 3.0]$. Figure \ref{fig:supp_camera_dist} shows the patch performance under this setting. 

For reference, we have included Figure \ref{fig:supp_camera_dist_renders}, which shows the size of our reference human mesh at many camera distances. The camera distances are with respect to 3D world coordinates. As we can see from Figure \ref{fig:supp_camera_dist}, the performance of our adversarial attack degrades significantly when the camera is translated towards or away from the human meshes. This motivates training under multiple camera distances, which is explored in the next section.

\subsection{Training with variable camera distance}
\label{sec:training_camera_dist}
\begin{figure}[h]
\centering
\includegraphics[width=1\linewidth]{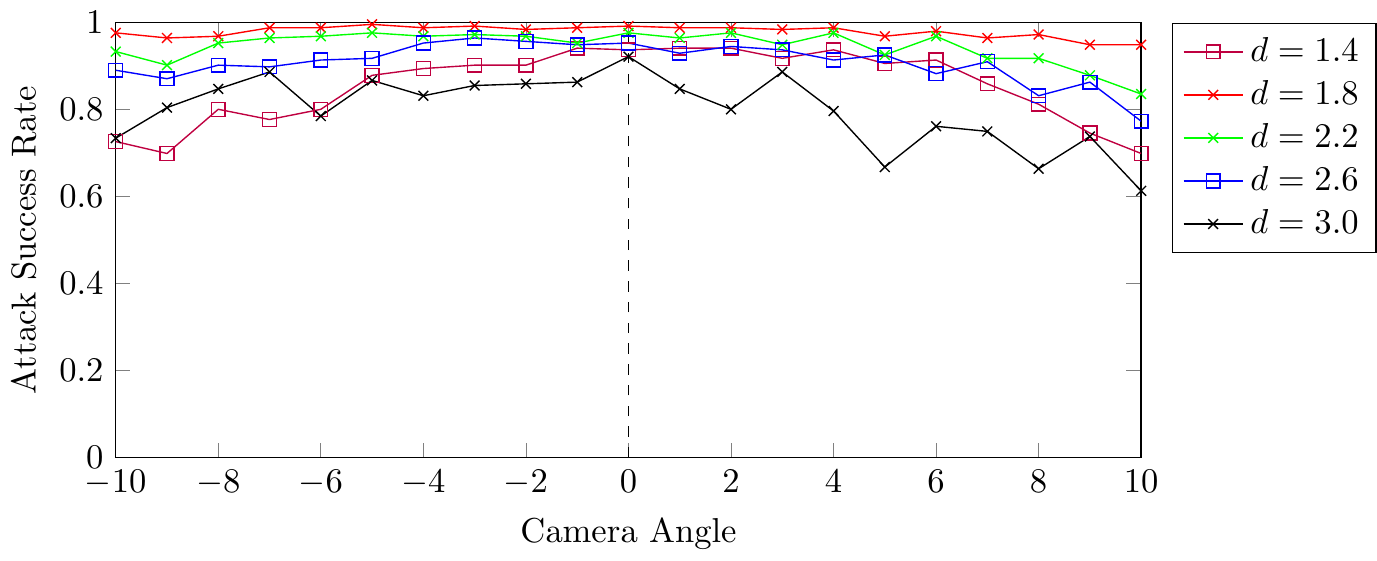}
\caption{The performance of a ``Chest + Thighs'' patch trained with the augmented camera distance described in Section \ref{sec:training_camera_dist}. We perform the same single-angle training and multi-angle testing described in Section 4.3.1. The patch is tested against various camera distances denoted by $d$. The green line (which is the default distance $d=2.2$) corresponds to the ``Chest + Thighs" setting back in Figure 6 in Section 4. \label{fig:supp_random_camera_dist}}
\vspace{-2mm}
\end{figure}

\begin{figure}[h]
\centering
\includegraphics[width=1\linewidth]{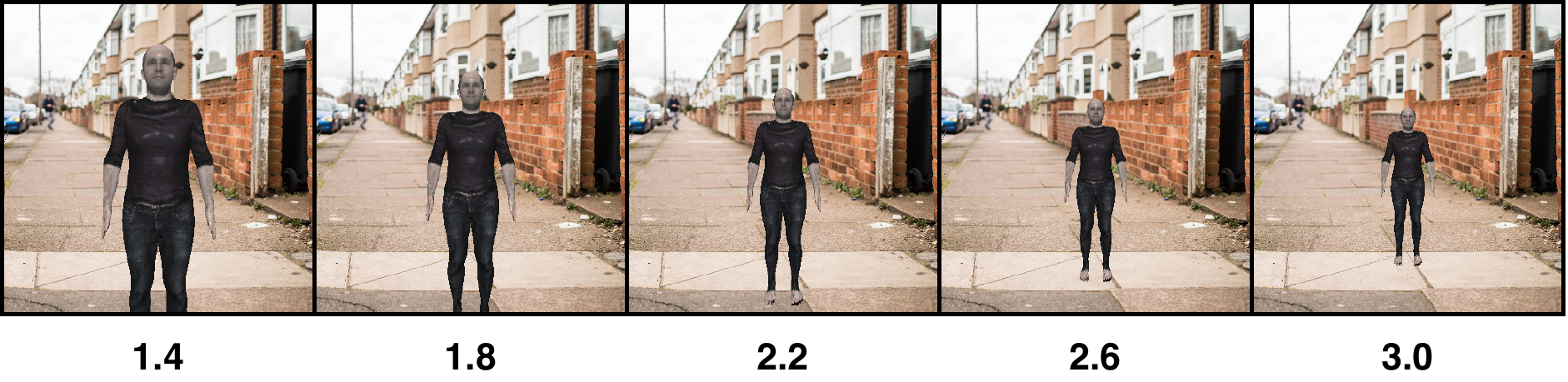}
\caption{A reference human mesh rendered at various camera distances. The closer distance shall reflect larger perceptive regions of attacking. }
\label{fig:supp_camera_dist_renders}
\end{figure}

In the second experiment, we slightly modified the training procedure to randomly perturb the camera distance for every batch. More specifically, we select a camera distance uniformly at random from $[1.4, 3.0]$ for each training batch. Our findings show that perturbing the camera distance during training produces stronger results for single-angle training. In Figure \ref{fig:supp_random_camera_dist}, we show the performance of this patch under varying camera distances. Not only is the performance much stronger than what is shown in Figure \ref{fig:supp_camera_dist}, but it even out-performs what is shown in Figure 6 of Section 4.3.2 (the ``Chest + Thighs'' patch). We can conclude that our patch is robust against perturbations in the angle and distance.

\section{Visualizing Total Variation Ablation Study}
\begin{figure*}[t]
\centering
\includegraphics[width=0.90\linewidth]{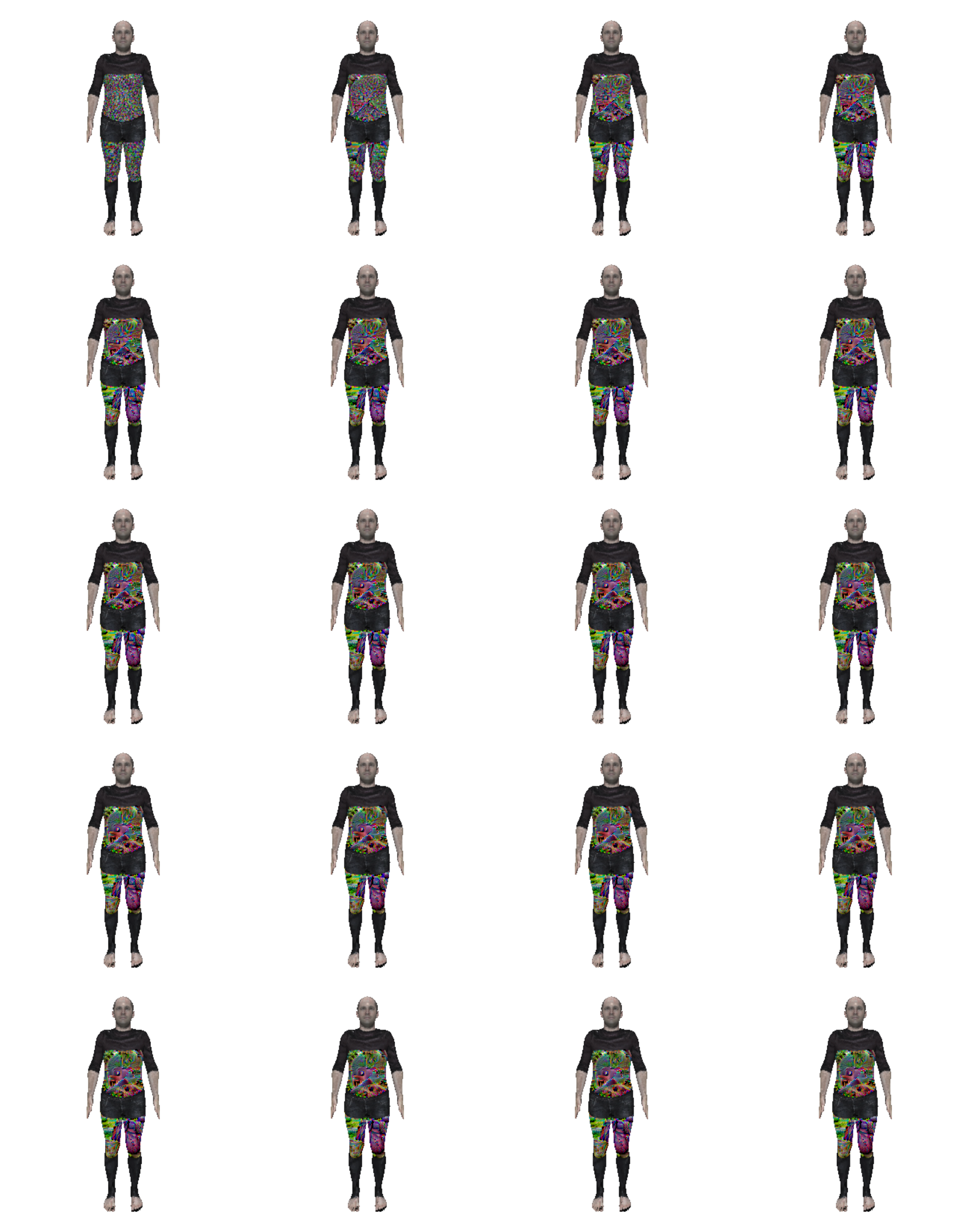}
\caption{The evolution of a ``Chest + Thighs'' patch trained against YoloV2 with $\lambda_{tv}=2.5$. From left to right, top to bottom, the images show the patch at every 5th epoch. Through our experiments, we found this to be the optimal total variation weight. Without the proper amount of smoothing, the patch will either be too detailed, which causes fluctuations in attacking performance, or it will contain too little detail, which diminishes the expressiveness of the patch. The patch displayed above displays an appropriate balance between the two.  \label{fig:supp_yolo_patches}}
\end{figure*}

\begin{figure*}[t]
\centering
\includegraphics[width=0.90\linewidth]{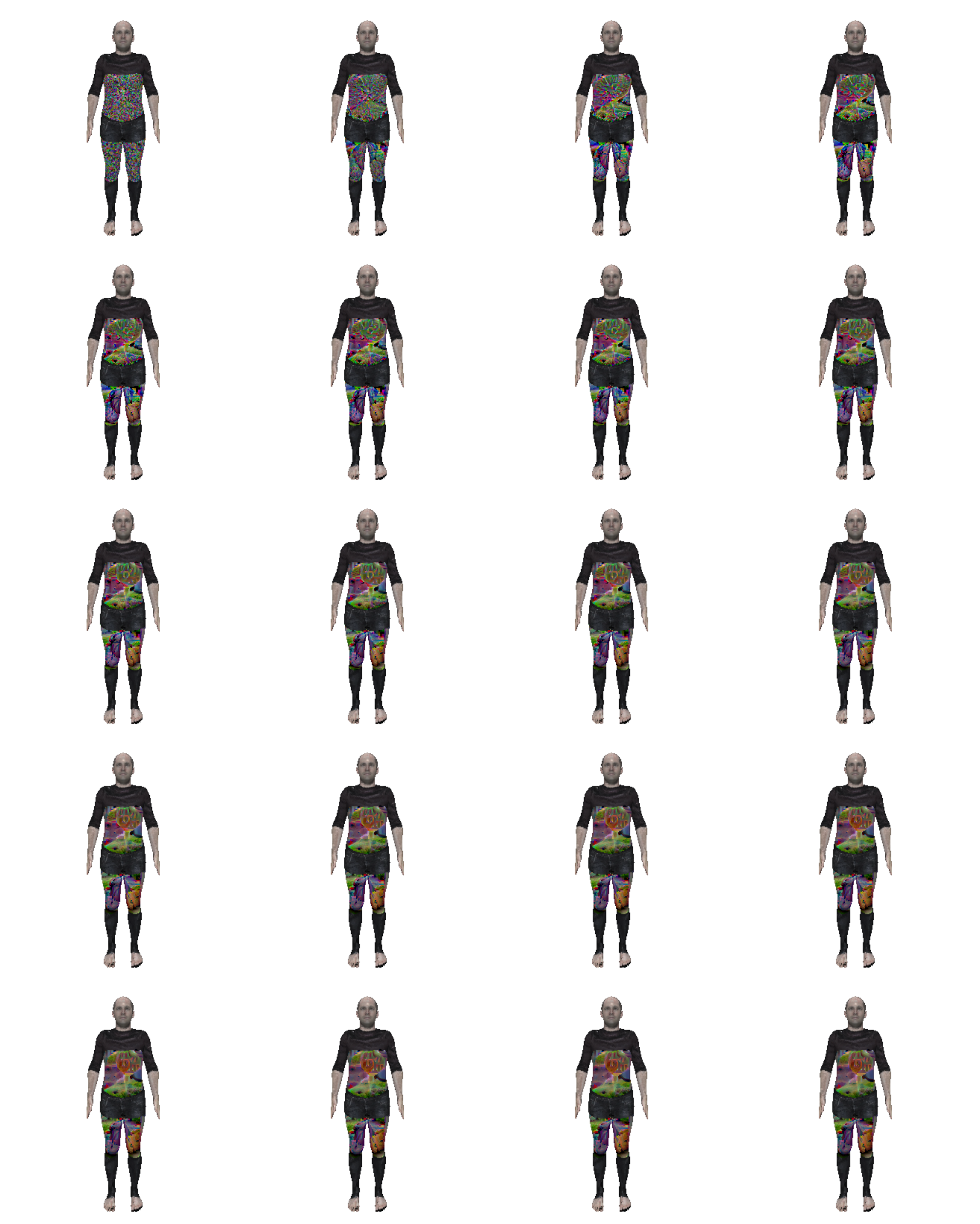}
\caption{The evolution of a ``Chest + Thighs'' patch trained against YoloV2 with $\lambda_{tv}=10$. From left to right, top to bottom, the images show the patch at every 5th epoch. Qualitatively, we can see that the resulting patch contains very few areas with minute detail. While the total variation loss is able to make the patch smooth, we observe a performance degradation when compared to the patch in Figure \ref{fig:supp_yolo_patches}. \label{fig:supp_yolo_patches_tv10}}
\end{figure*}

\begin{figure*}[t]
\centering
\includegraphics[width=0.90\linewidth]{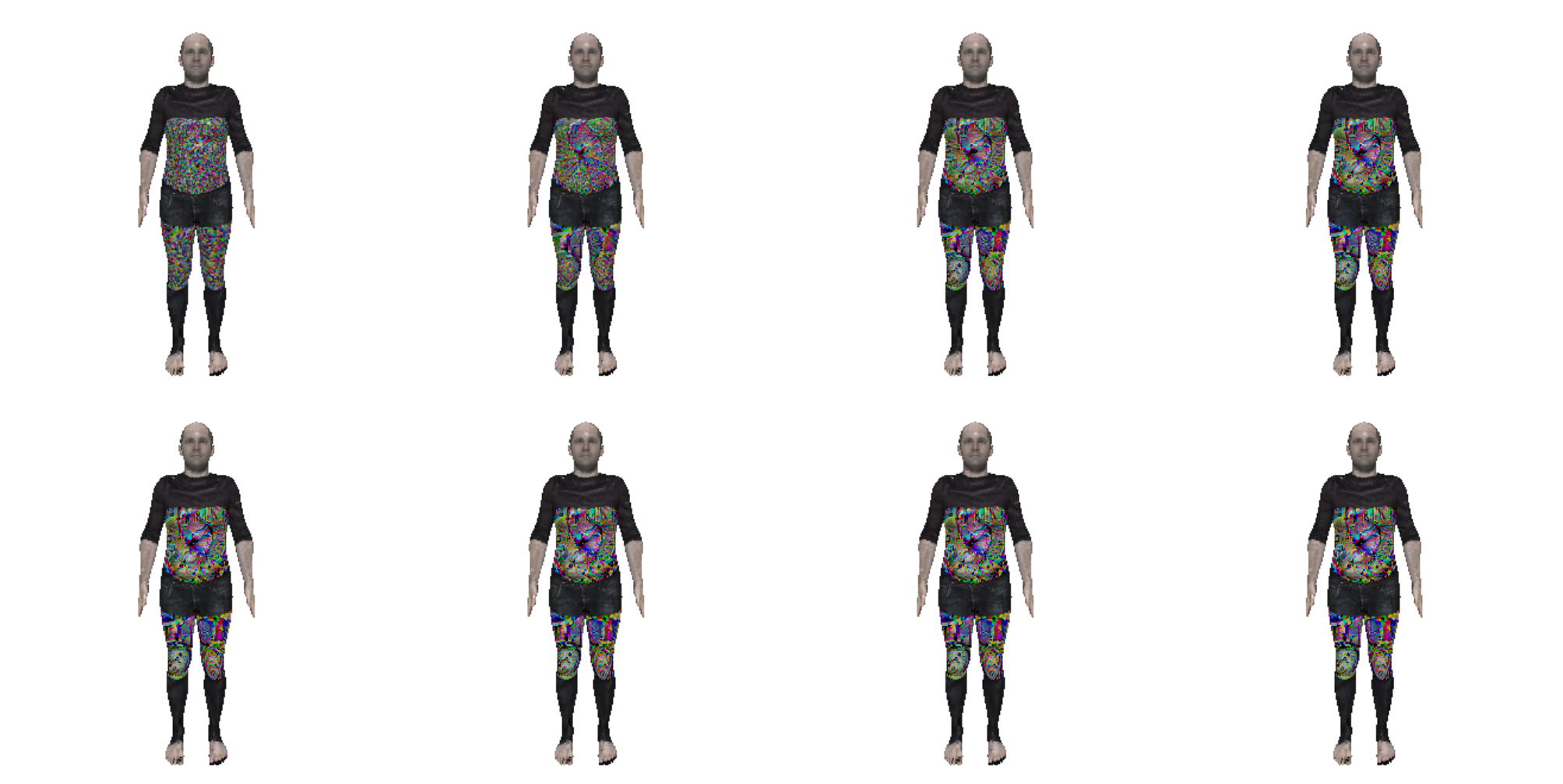}
\caption{The evolution of a ``Chest + Thighs'' patch trained against YoloV2 with $\lambda_{tv}=0.1$. From left to right, top to bottom, the images show the patch at every 5th epoch. As we can see, the resulting patch contains an extreme amount of noise, especially when compared to Figures \ref{fig:supp_yolo_patches_tv10} and \ref{fig:supp_yolo_patches}. The fine-grained detail in this patch causes it to appear visually inconsistent when rendered at different views. That is to say, its visible structure is dependent on viewing angle and distance. \label{fig:supp_yolo_patches_tv01}}
\end{figure*}

In Figures \ref{fig:supp_yolo_patches} and \ref{fig:supp_yolo_patches_tv10} we show the evolution of a ``Chest + Thighs'' patch trained against YoloV2 with $\lambda_{tv}=2.5$ and $\lambda_{tv}=10.0$ respectively. As we can see, the gradual formation of the patch is quite different. Qualitatively, the larger total variation weight removes much of the detail in the resulting patch. To showcase the importance of total variation, we show the same patch trained with $\lambda_{tv}=0.1$ in Figure \ref{fig:supp_yolo_patches_tv01}. The extreme amount of noise in this patch results in poor performance under perturbed camera angles.

\begin{figure*}[t]
\centering
\includegraphics[width=0.80\linewidth]{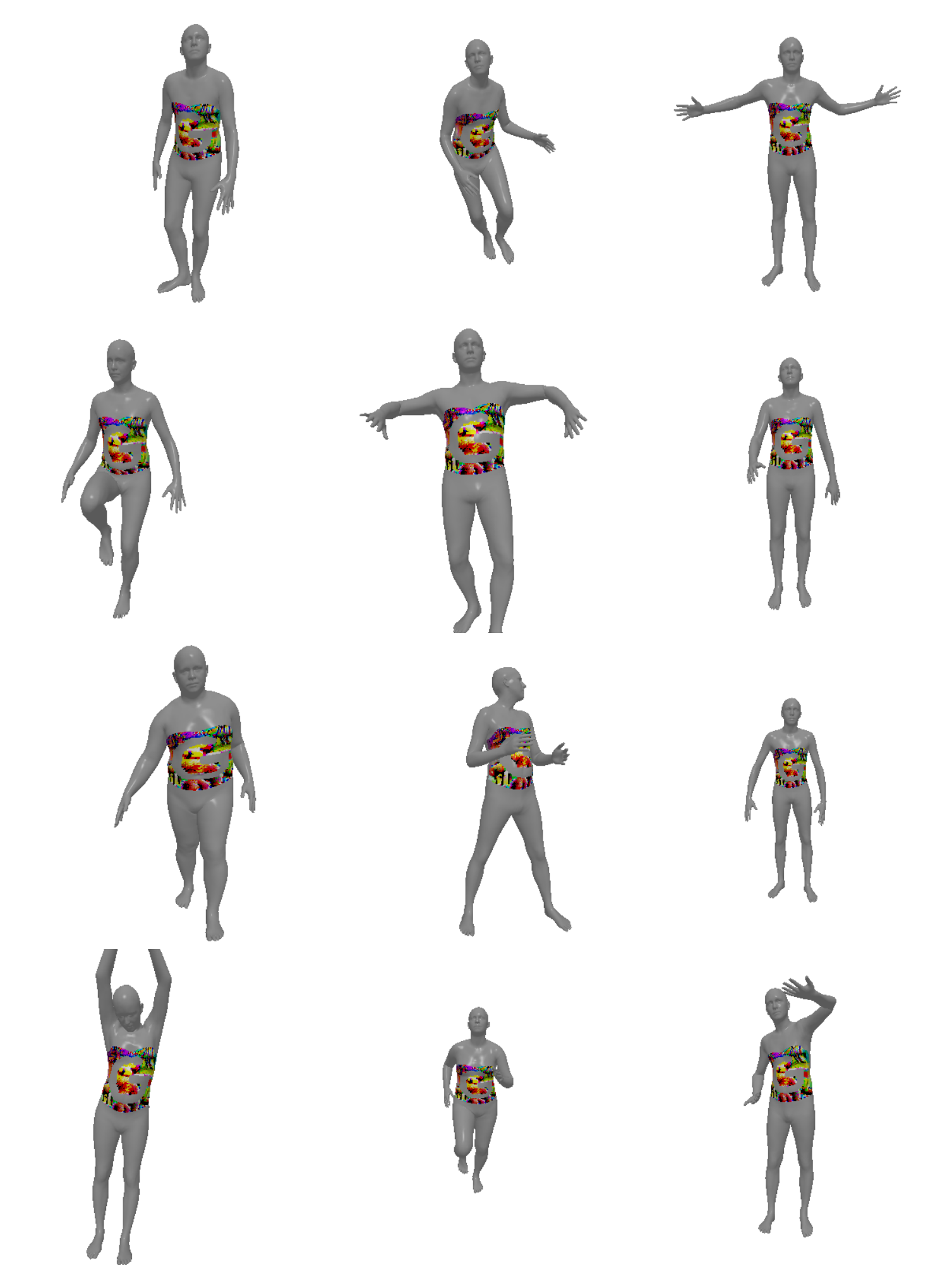}
\caption{A ``Letter G'' patch applied to human meshes in different poses. Due to topological consistency, the patch is transferable to each mesh and does not suffer from an egregious amount of distortion. Even in meshes with disparate body shapes, we can observe that the patch detail is largely preserved. This is crucial to our attack success rate, as the patch is able to produce a similar effect under many different scenarios.}
\end{figure*}

\begin{figure*}[t]
\centering
\includegraphics[width=0.80\linewidth]{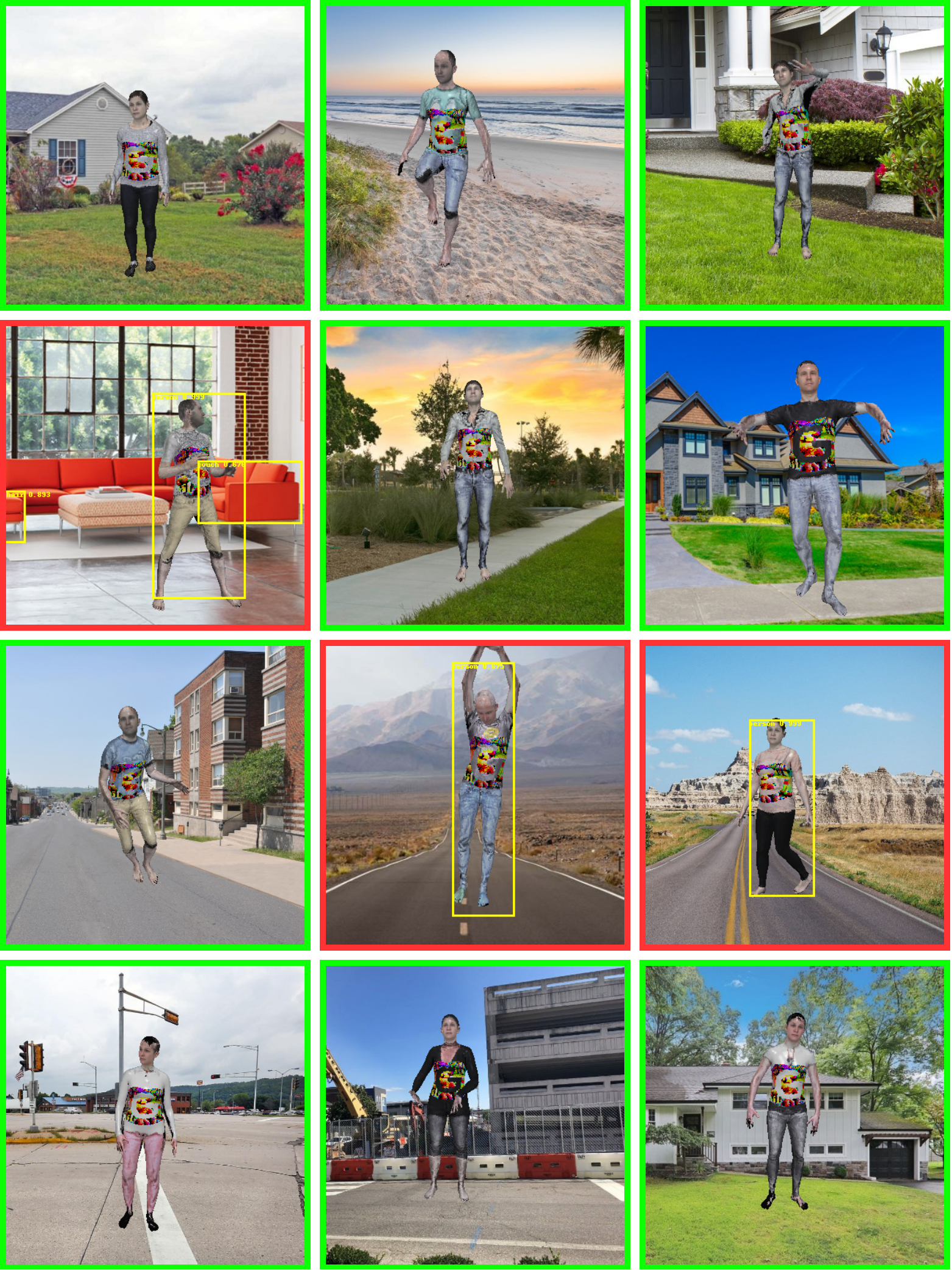}
\caption{A sample of the success and failure cases for a ``Letter G'' patch. Red borders indicate cases where the human mesh was identified by YoloV2. Green borders indicate instances of successful cloaking. From this limited sample, we can see that the failure cases include abnormal poses, perturbed camera angles, or patch occlusion.}
\end{figure*}

\end{document}